\pdfoutput=1

\documentclass[11pt]{article}

\usepackage[preprint]{acl}

\usepackage{times}
\usepackage{latexsym}
\usepackage{amsmath}
\usepackage{bm}
\usepackage{amsfonts}
\usepackage{amssymb}
\usepackage[most]{tcolorbox} 
\usepackage{xcolor} 
\usepackage{subcaption}

\usepackage[T1]{fontenc}

\usepackage[utf8]{inputenc}

\usepackage{enumitem}

\usepackage{microtype}

\usepackage{inconsolata}

\usepackage{graphicx}

\usepackage{booktabs}
\usepackage{adjustbox}
\usepackage{float}

\title{\modelname: A Large-Scale Generalist Diffusion Language Model}

\author{
 \textbf{Jaesung Tae\textsuperscript{1}\thanks{Co-first authors. Correspondence to \href{mailto:email@domain}{hamishi@allenai.org}.}},
 \textbf{Hamish Ivison\textsuperscript{2,3}}\footnotemark[1],
 \textbf{Sachin Kumar\textsuperscript{3,4}},
 \textbf{Arman Cohan\textsuperscript{1}}
\\
 \textsuperscript{1}Yale University,
 \hspace{6pt}\textsuperscript{2}University of Washington, \\
 \textsuperscript{3}Allen Institute for AI,
 \hspace{6pt}\textsuperscript{4}The Ohio State University \\
}

\newcommand{\modelname}{\textbf{TESS 2}}

\begin{document}
\maketitle
\begin{abstract}
We introduce \modelname, a general instruction-following diffusion language model that outperforms contemporary instruction-tuned diffusion models, as well as matches and sometimes exceeds strong autoregressive (AR) models. 
We train~\modelname~by first adapting an AR model via continued pretraining with the usual cross-entropy as diffusion loss, and then performing further instruction tuning. We find that adaptation training as well as the choice of the base model is crucial for training good instruction-following diffusion models. Furthermore, we propose reward guidance, a novel and modular inference-time guidance procedure to align model outputs without needing to train the underlying model.
Finally, we show that \modelname~further improves with increased inference-time compute, highlighting the utility of diffusion LMs in having fine-grained controllability over the amount of compute used at inference time.
Code and models are available at \url{https://github.com/hamishivi/tess-2}.
\end{abstract}

\section{Introduction}

Existing language models are predominantly trained using an autoregressive (AR) paradigm.
Prior works have suggested that AR models may have a number of limitations, especially in planning and self-correction~\citep{lin-etal-2021-limitations, pmlr-v235-bachmann24a, huang2024large}. While recent work has made large strides in improving AR model reasoning, this hinges on increasing test-time compute using specialized reinforcement learning-based training procedures~\citep{kumar2024traininglanguagemodelsselfcorrect,laskin2022incontextreinforcementlearningalgorithm} and costly long generations~\citep{deepseekai2025deepseekr1incentivizingreasoningcapability, min2024imitateexploreselfimprovereproduction}.

An alternative paradigm lies in~\textbf{diffusion language models}, which have shown promise in addressing some of the limitations of AR models~\citep{ye2024diffusion,zhang2023planner}. Diffusion models provide plug-and-play controllability at inference time~\cite{li2022diffusion}, are naturally capable of flexible generation such as infilling, and enable precise control of inference compute, more so than recent long chain-of-thought approaches. However, diffusion LMs have remained relatively small scale and focused on improving intrinsic metrics such as perplexity~\citep{lou2024discrete,gulrajani2023likelihoodbased}, rather than examining common downstream tasks used to evaluate AR LMs. This constrains their broader impact and practical applicability.

\begin{figure}[t]
    \centering
    \includegraphics[width=\linewidth]{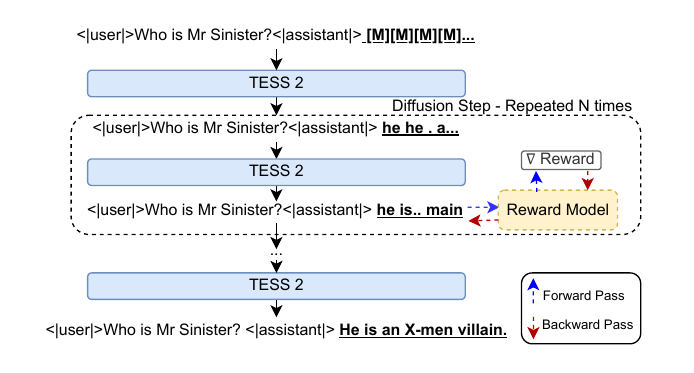}
    \caption{Overview of inference with \modelname. We provide a chat-template formatted query and iteratively denoise over the simplex space for a number of diffusion steps (typically 100). Optionally, we also incorporate reward guidance at each diffusion step by applying the gradient of the predicted reward to the intermediate logits, i.e., gradient ascent on reward.\vspace{-1mm}}
    \label{fig:high_level_view}
\end{figure}

In this work, we aim to close this gap by introducing~\modelname, a large-scale diffusion language model. We show that strong diffusion models can be trained by adapting from existing AR models, and that these models can be further instruction-tuned to serve as strong generalist models across varied downstream settings. We ablate a number of configurations and present a recipe for adapting existing AR LMs to diffusion models.
Our final model outperforms contemporary diffusion LMs and performs close to or exceeds AR baselines, showing that continuous diffusion models can serve as generalist instruction-tuned LMs. 

To the best of our knowledge, we are the first work to provide a \textbf{full recipe} for training generalist instruction-following fully non-autoregressive continuous diffusion models, with prior work only examining performance after task-specific finetuning~\citep{karimi-mahabadi-etal-2024-tess, han2024transferlearningtextdiffusion}, using semi-autoregressive setups and only testing chat settings~\citep{han2023ssd2}, or using a discrete diffusion approach instead of a continuous one~\citep{gong2024scalingdiffusionlm}. We also underscore the unique advantages of diffusion LMs over AR models: the scalability of inference-time compute with increased diffusion steps and the ability to supplement the backward diffusion process with reward models. We summarize our contributions.

\begin{enumerate}[itemsep=0mm]
    \item We propose a recipe for adapting LLMs into instruction-following diffusion LMs that \textbf{significantly outperform prior text diffusion models} and \textbf{perform similarly to or better than AR models} on QA and general instruction-following tasks.
    \item We \textbf{ablate components of the recipe and highlight salient elements}, such as the choice of the base model, where possible.
    \item We show that \textbf{our resulting model can further improve with increased test-time compute} by employing more diffusion steps.
    \item We introduce \textbf{reward guidance}, a novel technique for steering diffusion LMs to generate text aligned with user preferences, \textbf{enhancing downstream chat performance without any additional training}.
\end{enumerate}


\section{Related Work}

\paragraph{Diffusion Language Models} With the popularization of diffusion models in domains such as image~\citep{nichol2021improved}, audio~\citep{kong2020diffwave,tae2022editts,Shen2023NaturalSpeech2L}, video~\citep{ho2022video}, and text-to-image generation~\citep{saharia2022photorealistic, ramesh2022hierarchical}, there have been growing attempts to adapt and improve diffusion methods for language generation. One line of diffusion LMs proposes a diffusion process that adds Gaussian noise to word embeddings \cite{li2022diffusion,gulrajani2023likelihoodbased}. Another approach uses discrete diffusion based on categorical jump probabilities and continuous-time Markov chains~\citep{lou2024discrete,richemond2022categorical,gong2024scalingdiffusionlm,nie2025llada}, removing the need for mapping diffused embeddings to discrete inputs or outputs.

In contrast, simplex diffusion language models maintain the continuity of the diffusion process, instead aiming to learn discrete data by modelling over a continuous simplex. We build off TESS~\citep{karimi-mahabadi-etal-2024-tess}, which diffuses over the probability simplex and itself is a fully-non-autoregressive version of SSD-LM~\citep{han2022ssd}. While \citet{han-etal-2024-david} explored scaling SSD-LM for instruction-following, it remained semi-autoregressive and focused on a novel collaborative decoding strategy without releasing code, models, or data. In contrast, we show that fully non-autoregressive generation is possible at context lengths of up to 2048 tokens while only using publicly available data and model checkpoints. We also take a different approach from LLaDA~\cite{nie2025llada}, a contemporary work that focuses on pretraining from scratch at scale, and instead explore how we can leverage already pretrained AR models to reformulate them as diffusion LMs.


\paragraph{Adapting Diffusion Models} Concurrent and prior work has also examined instruction-tuning diffusion models and adapting them from existing AR models.
\citet{han2024transferlearningtextdiffusion} builds off SUNDAE~\citep{savinov2022stepunrolled} and shows that AR models can be successfully converted into diffusion models for further downstream finetuning. They focus on training diffusion checkpoints that can then be finetuned to perform specific tasks, as opposed to examining instruction tuning and general performance as done in this work. 

\citet{ye2023diffusionlanguagemodelsperform} examines scaling reparameterized discrete diffusion models~\citep{zheng2024reparameterizeddiscretediffusionmodel} by adapting XLM-R checkpoints~\citep{conneau-etal-2020-unsupervised}, but report that they are unable to adapt more modern LMs and find their models fall short on more complex reasoning tasks.

Concurrent to our work, \citet{gong2024scalingdiffusionlm} proposes DiffuLlama, an absorbing discrete diffusion model~\citep{austin2021structured} adapted from Llama~\cite{touvron2023llama2openfoundation}. However, they focus on discrete diffusion and pretraining without exploring instruction tuning. In comparison, we show that simpler continuous diffusion algorithms from SSD-LM and TESS can be straightforwardly translated to adapt diffusion LMs. We additionally explore instruction tuning adapted diffusion LMs and evaluate the model on common instruction tuning benchmarks, which better reflect real-life applications that require the ability to provide long-form generations that engage with user queries. 

\section{\modelname}

\subsection{Simplex Diffusion Language Models}

For the model architecture of \modelname, we largely follow TESS~\citep{karimi-mahabadi-etal-2024-tess}, a fully non-autoregressive diffusion model with self-conditioning. We provide a brief recap. 

\paragraph{Simplex-based Representation} Let $\mathcal{V}$ denote the vocabulary space. We map the index of each token to be generated $w \in \mathcal{V}$ to a $k$-logit simplex to produce $\mathbf{s}^w \in \{\pm k\}^{|\mathcal{V}|}$, whose $i$-th component satisfies
\begin{equation}
s^w_{(i)} = 
\begin{cases}
k, & \text{if}\quad i = w, \\
-k, & \text{otherwise},
\label{eq:scaled-simplex}
\end{cases}
\end{equation}
with a hyperparameter $k \in \mathbb{R}^+$. We then produce a probability simplex over $\mathcal{V}$ via $\mathbf{p}^w = \text{softmax}(\mathbf{s}^w)$. Finally, we compute the weighted sum of word embeddings to obtain a continuous embedding vector, $\mathbf{h}^w = \mathbf{E} \mathbf{p}^w$, where $\mathbf{E} \in \mathbb{R}^{d \times |\mathcal{V}|}$ is the word embedding matrix, $d$ denotes the size of the hidden dimension, and $\mathbf{h}^w \in \mathbb{R}^d$.

\paragraph{Timestep Embeddings}
We use a single linear layer without bias to produce timestep embeddings. These embeddings are directly added to the token embeddings before being fed into the first transformer block to inform the model of the current timestep in the diffusion process.

\paragraph{Forward Diffusion}
Let $\mathbf{w} = (w_1, \dots, w_L)$ be a sentence of $L$ tokens such that $w_i \in \mathcal{V}$, and $\mathbf{S}_0 = (\mathbf{s}^{w_1}, \dots, \mathbf{s}^{w_L}) \in \{\pm k\}^{L \times |\mathcal{V}|}$ be the $k$-logit simplex representation of $\mathbf{w}$. %
Following the standard Denoising Diffusion Probabilistic Models (DDPM) formulation~\citep{ho2020denoising}, we add noise to the $k$-logit simplex representation during training according to
\begin{align}
\mathbf{S}_t = \sqrt{\bar{\alpha}_t} \mathbf{S}_0 + \sqrt{1-\bar{\alpha}_t} \bm{\epsilon}_t, 
\label{eq:forward_diff}
\end{align}
where $t \in \{0, 1, \cdots, T\}$ is the timestep, $T \in \mathbb{N}$ is the total number of diffusion steps,  $\bm{\epsilon}_t \sim \mathcal{N}(0, k^2 \mathbf{I})$, and $\bar{\alpha}_t$ follows the cosine noise schedule~\cite{nichol2021improved}.

\paragraph{Training} We train the model by computing the usual cross-entropy loss between the ground-truth tokens $\mathbf{w}$ and the model prediction given a noisy logit simplex $\mathbf{S}_t$ at timestep $t$.
\begin{align}
\mathcal{L}
&= \mathbb{E}_{t,q(\mathbf{S}_0), q(\mathbf{S}_t | \mathbf{S}_0)} \left[ -\sum_{i = 1}^{L} \log p_{\bm{\theta}}(w_i | \mathbf{S}_t, t) \right].
\end{align}
This is in contrast to the usual mean squared error loss used in standard diffusion models, and is found to be stable in training simplex-based diffusion language models \cite{han2022ssd,karimi-mahabadi-etal-2024-tess}.

\paragraph{Sampling} During inference, we sample $\mathbf{S}_T$ from the prior $\mathcal{N}(0, k^2 \mathbf{I})$ and run the reverse process for $t = T,\dots, 1$ on the noisy $k$-logit simplex.
The reverse process can be approximated via
\begin{align}
\mathbf{S}_{t - 1} = \sqrt{\bar{\alpha}_{t-1}} \hat{\mathbf{S}}_{\bm{\theta}}(\mathbf{S}_t, t) + \sqrt{1-\bar{\alpha}_{t-1}} \bm{\epsilon}_t. \label{eq:rev_diff}
\end{align}
Thus, to reverse one step from $t$, we take the model prediction $\hat{\mathbf{S}}_{\bm{\theta}}$ as the hypothetical ground truth, then corrupt it by $(t - 1)$ timesteps.
To construct the model prediction, we project the logits predicted by the underlying encoder model via argmax as a pseudo-inverse of Eq.~\eqref{eq:scaled-simplex} to match the initial $k$-logit representation:
\begin{equation}
\hat{s}^{w}_{(i)} = 
\begin{cases} 
k, & \text{if\quad} i= \text{argmax}(\mathbf{s}^{w}), \\
-k, & \text{otherwise}.
\label{eq:scaled-simplex2}
\end{cases}
\end{equation}

\subsection{Adapting AR models to Diffusion}
Our approach to adapting AR models to diffusion consists of three key elements:
UL2 masking, label shifting, and full bidirectional attention.

\paragraph{UL2 Masking} 
Inspired by \citet{tay2023ul}, 
we train our model with a mixture of span infilling and prefix completion training objectives. For span infilling, we mask out random spans (with lengths randomly sampled from a pre-specified range) within the text and train the model to predict them. For prefix completion, we mask the last $n$ tokens, where $n$ is randomly chosen, and ask the model to predict the rest. The latter objective aligns with how we use the model for downstream tasks, but we found training with both objectives to work best for overall performance in pilot experiments. We provide further details in App.~\ref{app:further_training}.

\paragraph{Label Shifting}
Given some noised input, a typical diffusion LM predicts the less noised version of the same token at each position in the sequence. That is, given a noised simplex $\mathbf{S}_t = (\mathbf{s}_t^{w_1}, \dots, \mathbf{s}_t^{w_L})$, the model predicts $\mathbf{S}_{t - 1} = (\mathbf{s}_{t - 1}^{w_1}, \dots, \mathbf{s}_{t - 1}^{w_L})$. In our approach, we instead train the model to predict the token at the next index position to align with the next token prediction objective used for AR models, i.e., the model predicts $(\mathbf{s}_{t - 1}^{w_2}, \dots, \mathbf{s}_{t - 1}^{w_L + 1})$. This does not affect the validity of the denoising diffusion setup, but we find that it helps the model converge faster. Concurrent to our work, DiffuLlama~\cite{gong2024scalingdiffusionlm} also reported positive results with label shifting.

\paragraph{Full Bidirectional Attention} Finally, we disable causal masking and train our model with full bidirectional attention masks, as we found this led to better performance in pilot experiments. Fortunately, existing optimized model components such as Flash Attention~\citep{dao2022flashattention,dao2023flashattention2} are fully compatible with this change. This enables fast and efficient diffusion adaptation training. 
Concurrently, DiffuLlama~\cite{gong2024scalingdiffusionlm} also reported using full bidirectional attention.

\subsection{Instruction Tuning}

Once we have a diffusion-adapted model, we further put it through a stage of \textbf{instruction tuning}, in which we train the model on a smaller set of carefully curated instruction data. 
For this stage, we apply the same label shifting and full bidirectional attention mask as before, but switch from UL2 to only using a prefix-LM~\citep{j2018generating} objective for training. Specifically, we format our dataset to contain only single-turn instances, mask out the assistant response, and train the model to predict it. We also explored multi-turn training but observed marginal gains. We leave improved multi-turn training for future work.

\subsection{Reward-based Classifier Guidance}

Finally, we explore \textbf{reward guidance}, a novel method to improve model outputs without further training. We extend the guidance method proposed in \citet{han2022ssd} to be used with a scalar reward model. Intuitively, reward guidance works by adjusting the in-progress generation at each diffusion step using gradients from an off-the-shelf reward model, which is used to estimate the potential reward of the generation. This ultimately allows us to guide the generation process towards sequences with higher rewards.

Specifically, at each diffusion step, we take the output from the model at a given step, $\hat{\mathbf{S}}_{\bm{\theta}}$, and convert it to continuous token embeddings that we can then feed into an off-the-shelf reward model (with a matching tokenizer):
\begin{align}
    \mathbf{p}_t &= \text{softmax}(\hat{\mathbf{S}}_{\bm{\theta}}) \\
    \mathbf{c}_w & = \mathbf{E} \mathbf{p}_t,
\end{align}
where $\mathbf{E}$ is the embedding matrix of the classifier. We then pass $\mathbf{c}_w$ as input into the reward model, adding any required positional information before feeding it into the first hidden layer. This results in a predicted reward $R \in \mathbb{R}$ for the given timestep. Since we want to increase this, we perform \textbf{gradient ascent on the output reward}, computing the gradient of $\hat{\mathbf{S}}_{\bm{\theta}}$ with respect to the reward via backpropagation and then adding the resulting gradient to the model prediction by
\begin{align}
    \hat{\mathbf{S}}_{\bm{\theta}} := \hat{\mathbf{S}}_{\bm{\theta}} + \eta \cdot \nabla_{\mathbf{\theta}} R,
    \label{eq:reward_guidance}
\end{align}
where $\eta$ is the guidance coefficient that controls how much guidance to apply at each timestep.
Note that this is an inference-time algorithm that is applied at each timestep during the backward process; we do not apply any guidance during training.

\section{Experiments}

\subsection{Setup}
\label{sec:setup}
We evaluate our model in three settings: directly after diffusion adaptive pretraining, after instruction tuning, and with reward guidance applied.

\paragraph{Diffusion Adaptation} For diffusion adaptation training, we train the model on Dolma 1.7 \cite{soldaini-etal-2024-dolma}, an open pretraining dataset, for up to 200,000 train steps, or approximately 45 billion tokens.\footnote{This is roughly in line with prior work;~\citet{han2023ssd2} trained on 38B tokens from C4.} We use a constant learning rate of $1\times10^{-5}$ with a linear warmup of 5000 train steps, and a batch size of 112 samples.\footnote{While it is common to decay the learning rate, we use a constant LR for ease of experimentation and restarts.} We adapt the Mistral-7B-v0.1 model~\citep{jiang2023mistral7b}, based on model ablations explored in Sec.~\ref{sec:continued_pretraining}. We use a simplex value of 5 and self-conditioning as done in \citet{karimi-mahabadi-etal-2024-tess}. We provide additional details in App.~\ref{app:further_training}.

\paragraph{Instruction Tuning} For instruction tuning, we use the Tulu 2 SFT mixture~\citep{ivison2023camelschangingclimateenhancing}, which consists of roughly 326k samples. We finetune for 3 epochs with a learning rate of $1\times10^{-5}$ with linear warmup for the first 3\% of training steps and linear cooldown for the rest of training, which we found worked best in pilot experiments.

\paragraph{Reward Guidance} For reward guidance, we train a standard reward model on human preferences using Mistral-7B-v0.1 and the data and hyperparameters provided by \citet{dong2023raft, xiong2023gibbs}. We provide training details in App.~\ref{app:rm_training}.

\paragraph{Adaptation Evaluation} Following \citet{han2022ssd, han2023ssd2, richemond2022categorical}, we evaluate our adaptation using perplexity, average distinct $n$-grams in the output samples (d-1/2/3/4), unigram (per-token) entropy, and Mauve score~\citep{pillutla-etal:mauve:neurips2021}. Perplexity is measured using a held-out model,\footnote{Specifically, GPT-NeoX 1.3B~\citep{gpt-neo}.} following \citet{han2022ssd,han2023ssd2}.

\paragraph{Downstream Evaluation} 
In addition to the adaptation evaluation, we evaluate our models on a diverse set of downstream tasks: (1) \textbf{AlpacaEval}~\citep{alpaca_eval}, which tests general instruction following capabilities; (2) \textbf{GSM8k}~\citep{cobbe2021gsm8k}, which tests simple math word problems; (3) \textbf{SQuAD}~\citep{rajpurkar-etal-2016-squad} (with context), which tests general question-answering abilities; (4) \textbf{TriviaQA}~\cite{joshi-etal-2017-triviaqa}, a reading comprehension dataset; (5) \textbf{IFEval}~\citep{zhou2023instruction}, which tests exact instruction-following; and (6) \textbf{Big Bench Hard (BBH)}~\citep{suzgun2022challenging}, which tests general reasoning. For GSM8k, we additionally test the performance of our model after finetuning on the GSM8k symbolic data provided by \citet{ye2024diffusion}, following DiffuLLaMA \cite{gong2024scalingdiffusionlm}. We provide further details in App.~\ref{app:eval_details}.

\subsection{Diffusion Adaptation}
\label{sec:continued_pretraining}

\begin{table}[!ht]
\centering
\adjustbox{width=\linewidth}{
\begin{tabular}{@{}lcccccc@{}}
\toprule
 & \textbf{Rand.} & \textbf{RoBERTa} & \textbf{Llama 2} & \textbf{Llama 3} & \textbf{Mistral} \\ \midrule
Perp. & 54.4 & \textbf{20.2} & 3619.2 & 880.4 & 24.3 \\
d-1 & 0.55 &  0.36 & 0.94 & \textbf{0.97} & 0.62 \\
d-2 & 0.82 & 0.59 & \textbf{0.99} & \textbf{0.99} & 0.87 \\
d-3 & 0.91 & 0.69 & \textbf{0.98} & \textbf{0.98} & 0.93 \\
d-4 & 0.94 & 0.75 & \textbf{0.98} & \textbf{0.98} & 0.94 \\
Mauve & 0.92 & 0.93 & 0.01 & 0.93 & \textbf{0.95} \\
Entropy & 5.7 & 4.8 & 7.7 & \textbf{7.8} & 6.3 \\ \bottomrule
\end{tabular}}
\caption{Performance of diffusion adaptation after 35,000 steps of pretraining over varied base models. In all cases except perplexity, higher is better. Rand. inidicates initializing from scratch.}
\label{tab:pretraining_model}
\end{table}

\begin{figure}[t]
    \centering
    \includegraphics[width=\linewidth]{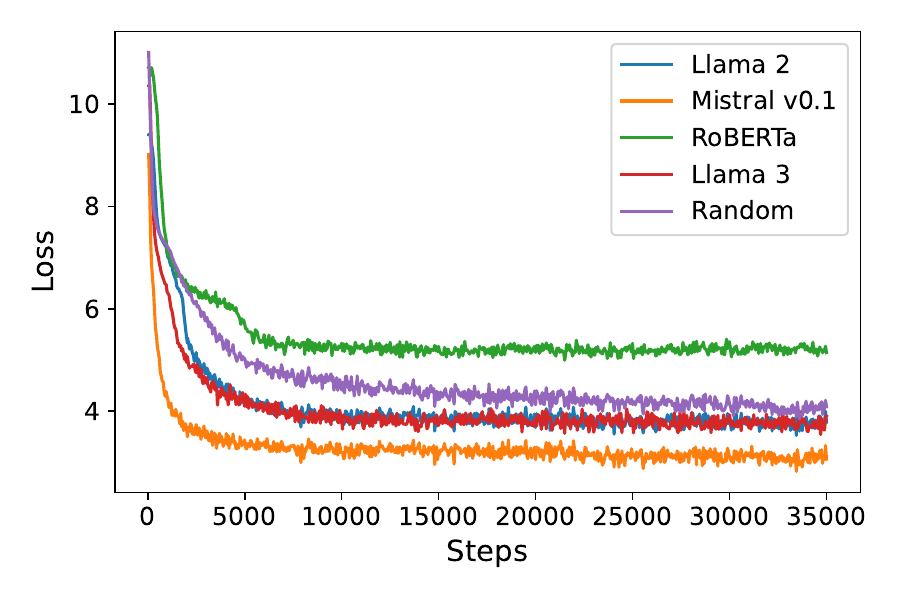}
    \caption{Train loss during adaptation training when adapting from different models. We find Mistral achieves the lowest overall loss during training, even compared to newer LMs such as Llama 3.}
    \label{fig:pretraining_loss}
\end{figure}

While TESS used a RoBERTa model~\citep{liu2019robertarobustlyoptimizedbert} as its backbone, recently there has been a wave of increasingly larger and more powerful open-weight pretrained models~\citep{jiang2023mistral7b,touvron2023llama2openfoundation,grattafiori2024llama3herdmodels}. However, unlike RoBERTa, these models are typically decoder-only causal LMs, which makes it unclear how well they transfer to the full bidirectional attention setup used for diffusion LMs. This additional layer of complexity has been a challenge in adapting pretrained LLMs to diffusion.

We investigate this by conducting diffusion adaptation training on 4 different base models for 35,000 steps (roughly 2B tokens) with Dolma 1.7: RoBERTa~\citep{liu2019robertarobustlyoptimizedbert}, Llama 2~\citep{touvron2023llama2openfoundation}, Llama 3.0~\citep{grattafiori2024llama3herdmodels}, and Mistral base v0.1~\citep{jiang2023mistral7b}. We additionally try pretraining from scratch using a randomly initialized Mistral v0.1. We use a sequence length of 512 due to RoBERTa's only supporting up to that length. We evaluate these models on a random sample of 512 samples from the C4 validation set~\citep{2020t5}, using the metrics described in Sec.~\ref{sec:setup}.
We show the results in Table~\ref{tab:pretraining_model}. We enumerate our findings below.

\paragraph{Llama models do not adapt well.} Llama models were unable to generate coherent text with the small number of pretraining steps. While we find the models perform better with more pretraining, they additionally display higher loss values throughout training compared to Mistral or RoBERTa, as seen in Figure~\ref{fig:pretraining_loss}. This also occurs for Llama 3.0, despite it being a stronger base model than Mistral v0.1 in AR settings. We suggest this is due to the use of bidirectional attention during training, which results in a large change in the model's internal hidden states.\footnote{Concurrent work attempts to slowly convert the causal mask to a bidirectional one when adapting Llama models, but find it provides minimal improvements~\citep{gong2024scalingdiffusionlm}.} We additionally experimented with maintaining the causal attention mask during pilot experiments and found it still yielded poorer training curves compared to Mistral, likely due to the fact that the causal mask removes a core benefit of diffusion LMs: full bidirectional attention flows across the entire input and sequence to be generated. In contrast, Mistral likely behaves better due to potentially being pretrained with a prefix-LM objective with partial bidirectional attention; its ability to handle sudden changes in its attention mask has also been noted in prior work~\citep{behnamghader2024llmvec}, which hypothesized that Mistral may have been trained with a prefix-LM objective. 

\paragraph{Adapting Mistral results in coherent and diverse generations.} Additionally, we find that Mistral performs better than Llama, achieving much lower perplexity and generating coherent text. While its perplexity is higher than RoBERTa, Mistral also displays significantly better distinct $n$-grams and Mauve, indicating diverse, high quality generations. Furthermore, Mistral is able to natively handle much longer sequences than RoBERTa, which is limited to 512-length sequences due to its absolute positional embeddings. Finally, the Mistral adaptation produces lower loss than training from scratch, suggesting that the diffusion-adapted model is able to make use of the knowledge and skills attained through AR pretraining.

\begin{table}[!ht]
\centering
\begin{tabular}{@{}lccc@{}}
\toprule
\textbf{Ctx. Len.} & 512 & 2048 & 2048 \\
\textbf{Train Steps} & 35,000 & 10,000 & 200,000 \\ \midrule
Perp. & 24.3 & 12.2 & \textbf{11.3} \\
d-1 & \textbf{0.62} & 0.58 & 0.59 \\
d-2 & \textbf{0.87} & 0.84 & 0.85 \\
d-3 & \textbf{0.93} & 0.91 & 0.91 \\
d-4 & \textbf{0.94} & 0.93 & 0.93 \\
Mauve & 0.95 & \textbf{0.97} & 0.92 \\
Entropy & 6.31 & 6.21 & \textbf{6.34} \\ \bottomrule
\end{tabular}
\caption{Intrinsic pretraining metrics when pretraining Mistral 7b v0.1 base with varying context lengths and pretraining steps. We find that using a longer context length and training for longer improves performance.}
\label{tab:ctx_len_exp}
\end{table}

Given these results and our limited compute budget, we chose Mistral as the base for future experiments. However, we emphasize this does not imply that Llama is inherently unsuitable for our approach: the Llama models are far from full convergence given that we saw a measurable improvement in performance at least up to 200,000 steps of diffusion adaptation with our final model, as shown in Figure~\ref{fig:adaptation_steps}. Instead, we believe that Llama could also benefit from our proposed recipe, though achieving comparable results may require additional compute for adaptation training.

As a next step, we further scale the training of \modelname~by (1) extending the sequence length to 2048 tokens, and (2) training up to 200,000 training steps. Results are shown in Table~\ref{tab:ctx_len_exp}. We find that increasing the context length instantly improves performance, with the model achieving much lower perplexity and a higher mauve score while remaining reasonably diverse in terms of generations. Additionally, we observe that training for longer further reduces perplexity while minimally affecting diversity. We also find that longer adaptation training results in improved downstream performance after instruction tuning (see Sec.~\ref{sec:pretrain_len_downstream}).

\subsection{Instruction Tuning}

\begin{table*}[!ht]
\centering
\adjustbox{max width=\textwidth}{
\begin{tabular}{lcccccc|c}
\toprule
\textbf{Model} & \textbf{Alp. Eval} & \textbf{SQuAD} & \textbf{TriviaQA} & \textbf{IFEval} & \textbf{BBH} & \textbf{GSM8k} & \textbf{GSM8k (ft)} \\
\midrule
Mistral v0.1 AR & \underline{77.1} & \underline{86.0} & 50.4 & 36.8 & \underline{43.3} & 52.5 & 51.2 \\
\quad w/ cont. pretrain & 73.6 & 71.4 & 20.5 & 32.0 & 32.2 & 40.7 & - \\ 
Mistral v0.3 AR & 63.3 & 48.9 & 36.7 & \underline{62.7} & 38.4 & \underline{54.7} & 45.6 \\
\midrule
DiffuLlama & 0.2 & 34.9 & 19.7 & 14.4 & 1.9 & 0.0 & 63.1$*$ \\
Flan-XLM-R-D XXL & 0.0 & 41.0 & 0.0 & 12.2 & 1.2 & 3.0 & 12.8$^\dagger$ \\ \midrule
\modelname~\textbf{(ours) v0.1} & \textbf{63.1} & \textbf{85.4} & 49.3 & 30.5 & 8.4 & 14.5 & \underline{\textbf{66.6}} \\
\quad w/o diffusion adaptation & 0 & 3.6 & 2.9 & 12.9 & 1.0 & 1.1 & 0.2 \\ 
\modelname~\textbf{(ours) v0.3} & 62.2 & 84.8 & \textbf{\underline{53.8}} & \textbf{54.6} & \textbf{10.8} & \textbf{36.5} & 59.2 \\
\bottomrule
\end{tabular}}
\caption{Performance of various models on downstream tasks after undergoing instruction tuning, including DiffuLlama~\citep{gong2024scalingdiffusionlm} and Flan-XLR-R-D~\citep{ye2023diffusionlanguagemodelsperform}. `GSM8k (ft)' refers to finetuning on the augmented GSM8k symbolic dataset proposed by \citet{ye2024diffusion} and then evaluating on GSM8k, following \citet{gong2024scalingdiffusionlm}.
We \textbf{bold} the best result from a diffusion LM and \underline{underline} the best overall result.
* Number from \citet{gong2024scalingdiffusionlm}.
$\dagger$ Number from \citet{ye2024diffusion} when finetuning on a different distilled GSM8k dataset.}
\label{tab:instruction_performance}
\end{table*}

We now explore how well our adapted model performs after instruction tuning. We perform diffusion adaptation on Mistral v0.1 for 200,000 steps and then further instruction tune on the Tulu 2 SFT mixture~\citep{ivison2023camelschangingclimateenhancing}, a popular instruction tuning dataset.
We additionally finetune models on the augmented GSM8k symbolic set from \citet{ye2024diffusion} and evaluate on GSM8k, following \citet{gong2024scalingdiffusionlm}.
Finally, we also explore updating our recipe with newer models and datasets, using Mistral v0.3 and the Tulu 3 SFT mixture, following the same adaptation strategy and hyperparameters as the Mistral v0.1 and Tulu 2 setup. We compare~\modelname~with (1) the same Mistral models AR-finetuned on the Tulu 2/3 data (2 for v0.1, 3 for v0.3); (2) prior diffusion LMs of similar size to \modelname, DiffuLlama~\cite{gong2024scalingdiffusionlm} and Flan-XLM-R-D XXL~\cite{ye2023diffusionlanguagemodelsperform}. We also train Mistral v0.1 on the same number of tokens as \modelname~and then finetune on Tulu 2 (w/ cont. pretrain). We provide more details in App.~\ref{app:baselines}.
We use 100 diffusion steps at inference time for all diffusion models.
We show the results in Table~\ref{tab:instruction_performance}, denoting \modelname~\textbf{v0.1} as our model trained using the Mistral v0.1 model and Tulu 2 data, and \modelname~\textbf{v0.3} as our model train using the Mistral v0.3 model and Tulu 3 data.
We list our findings below.

\paragraph{\modelname~outperforms other diffusion models.} \modelname~v0.3 outperforms all other diffusion LMs across all evaluations (with \modelname~v0.1 often a close second). This is likely due to the simplicity of the simplex diffusion setup with cross-entropy, as well as the careful selection of a stronger and more compatible base model (Sec.~\ref{sec:continued_pretraining}). 
This results in better generations, even for baselines trained on more tokens: DiffuLlama is trained on 65B tokens, compared to only 45B for \modelname.

\paragraph{\modelname~performs well on general QA tasks but poorly on reasoning.} \modelname~performs close to or better than AR counterparts on AlpacaEval, SQuAD, and TriviaQA, showing that it can perform QA and produce coherent long-form generations (as tested by AlpacaEval) as well as, if not better than, AR models.
However, we find that \modelname~still lags behind Mistral AR for reasoning-centric tasks such as BBH and GSM8k. While we find that increasing the amount of inference-time compute used can reduce this gap (Sec.~\ref{sec:inference_time_compute}), it remains.

We note that Mistral was pretrained on an unknown data mixture, and continued pretraining on Dolma seems to hurt Mistral’s mathematical reasoning abilities. This gap in data quality was similarly reported in OLMo 2~\cite{olmo20252olmo2furious}, which found that OLMo 1 scored 27.7\% on GSM8k, while Mistral achieved 40.1\%. We speculate that Mistral’s pretraining data had a higher mix of quality mathematical datasets than Dolma, suggesting that training~\modelname~on a better data mixture may help close this gap.


We further investigate this performance gap by examining generations from the different diffusion models and AR models. Qualitatively, we observe that \modelname~is indeed able to produce coherent generations, but often makes basic mistakes in its reasoning, resulting in low exact match scores. Other diffusion models often fail to even produce coherent generations, explaining their overall low performance.
We show these generations in App.~\ref{app:reasoning_generations_app}.


\paragraph{\modelname~outperforms AR when large amounts of domain-specific data are available.} However, when finetuning on the GSM8k symbolic dataset, surprisingly we find that \modelname~benefits significantly, with both \modelname~variants outperforming both Mistral AR variants.
This suggests that diffusion models can achieve performance similar to AR models when using more data, and suggests that further tailoring instruction-tuning mixtures for diffusion models (e.g., adding more math and reasoning data) is a promising direction.


\subsection{Stepping Up with Reward Guidance}

\begin{figure*}[t]
    \centering
    \begin{subfigure}[t]{0.32\textwidth}
        \centering
        \includegraphics[width=\linewidth]{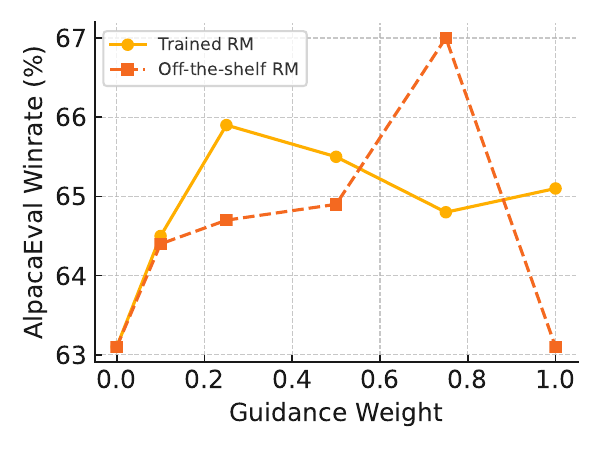}
        \caption{AlpacaEval performance against reward guidance weight. Increasing guidance weight initially improves, and then degrades performance.}
        \label{fig:reward_guidance}
    \end{subfigure}
    \hfill
    \begin{subfigure}[t]{0.32\textwidth}
        \centering
        \includegraphics[width=\linewidth]{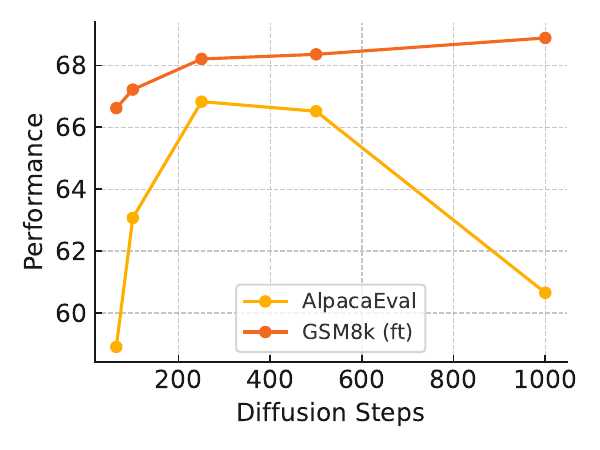}
        \caption{AlpacaEval and GSM8k performance using increasing diffusion steps at inference time. Performance increases with number of steps up to a point.}
        \label{fig:diff_steps_perf}
    \end{subfigure}
    \hfill
    \begin{subfigure}[t]{0.32\textwidth}
        \centering
        \includegraphics[width=\linewidth]{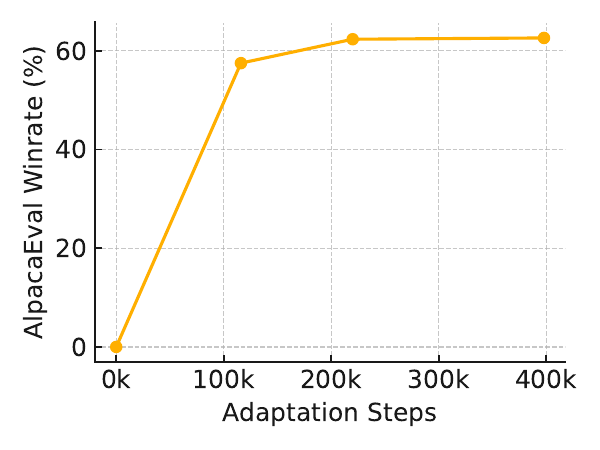}
        \caption{AlpacaEval winrate against number of diffusion adaptation steps. Going up to 200k steps provides significant improvements.}
        \label{fig:adaptation_steps}
    \end{subfigure}
    \caption{Analysis Experiments on \modelname.}
\end{figure*}

\label{sec:reward_guidance}

Finally, we explore applying reward guidance to \modelname~v0.1. We vary the reward guidance weight ($\eta$ in Eq.~\eqref{eq:reward_guidance}) when generating responses for AlpacaEval, and plot the result in Table~\ref{fig:reward_guidance}. 

\paragraph{Reward guidance helps.} We find that using reward guidance can further improve AlpacaEval performance, providing a 3-point gain without requiring further training when using a guidance weight of 0.25. We find that using higher guidance weights can still provide a small boost, while using significantly higher values results in nonsensical generations (consisting mostly of dashes), as the generation process becomes dominated by the gradients from the reward model. These degenerate outputs still have high reward, but are no longer semantically meaningful.\footnote{This is similar to ``reward-hacking" in RLHF, where models learn how to generate meaningless sequences that nonetheless achieve high reward~\citep{bai2022traininghelpfulharmlessassistant}.} As such, reward guidance provides a way to further improve and guide model generations without additional retraining.

\paragraph{Reward guidance is generally applicable across RMs.} Moreover, we find that reward guidance is generally applicable across different reward models. We applied reward guidance on AlpacaEval using (1) an RM trained using \texttt{open-instruct}~\footnote{\url{https://github.com/allenai/open-instruct}}, a widely used open-source reward model training recipe, and (2) \texttt{weqweasdas/RM-Mistral-7B}~\footnote{\url{https://huggingface.co/weqweasdas/RM-Mistral-7B}}, a highly-ranked Mistral RM on RewardBench~\cite{lambert2024rewardbenchevaluatingrewardmodels}. As seen in Figure~\ref{fig:reward_guidance}, we find that reward guidance boosts AlpacaEval performance by up to 3 points using our trained RM and 4 points with the off-the-shelf model, suggesting that reward guidance generalizes across RM.

\section{Analysis}

\subsection{Scaling Inference-time Compute}
\label{sec:inference_time_compute}

One advantage of diffusion models is that inference-time compute can be directly configured by varying the number of backward steps. Figure~\ref{fig:diff_steps_perf} shows that more steps generally yield better performance, with GSM8k performance demonstrating a consistent upward trend up until 1000 steps. Interestingly, in AlpacaEval, the score decreases significantly after 500 steps. We find that this is due to increased model repetitions with higher diffusion steps. This results in lower perplexity but less diverse text, which is penalized by the LM judge.


\subsection{Adaptation Training Steps and Downstream Performance}
\label{sec:pretrain_len_downstream}

While we showed perplexity improvements from increased adaptation steps in Sec.~\ref{sec:continued_pretraining}, this may not necessarily translate to downstream improvements. To further investigate, we additionally examine downstream AlpacaEval performance against the number of training steps. We run diffusion adaptation training on Mistral v0.1 using the same hyperparameter settings as \modelname, except using a shorter sequence length (1024) to reduce compute costs, and evaluate checkpoints throughout training up to 400k steps. We plot our results in Figure~\ref{fig:adaptation_steps}. As seen, we find that performance drastically improves up to 200k adaptation steps, and somewhat plateaus after. As such, we chose to use 200k steps of adaptation for our final \modelname~run.

\subsection{Evolution of Diffusion Predictions}
\label{sec:predictions_evolve}

We further investigate how diffusion predictions evolve over diffusion timesteps by plotting the confidence of \modelname~v0.1 as measured by the top-1 token probability over time. We find that the sequence is largely determined about 60\% of the way through the diffusion steps, and interestingly the later portion of the generation is sometimes generated first. This suggests that early stopping of the generation may be a promising method of reducing inference compute costs, as observed in prior work~\citep{han2023ssd2}.
We visualize the predictions in Figure~\ref{fig:confidence_over_steps} in App.~\ref{app:confidence_over_steps}.

\subsection{Sampling Speed}
\label{sec:sampling_speed}

Diffusion models are notoriously slow due to the number of backward steps required. To quantify the sampling cost of our model, we ran the model on AlpacaEval with naive generation via Hugging Face \texttt{transformers}~\cite{wolf-etal2020transformers} with a maximum output length of 2048, batch size of 8, and Flash Attention 2~\cite{dao2023flashattention2} enabled. Surprisingly, we find that it takes approximately 480 seconds per batch for the AR model and 77 seconds per batch for the diffusion model. This occurs because the diffusion model uses 100 forward passes per output even if the output is over 100 tokens in length, while the AR generation loop has to perform up to 2048 forward passes. We provide a basic sketch on the asymptotic complexity of diffusion and AR models in App.~\ref{app:sampling_speed}. Given these analyses, we speculate that this discrepancy will further widen in favor of diffusion LMs as we move to longer sequence regimes, consistent with the findings from~\citet{karimi-mahabadi-etal-2024-tess}.

\section{Conclusion}

We present \modelname, a large-scale generalist instruction-following diffusion language model that outperforms state-of-the-art diffusion models and achieves parity with comparable AR models on a number of tasks.
We propose an effective recipe for adapting frontier open-weight LMs to diffusion. Our recipe involves a combination of masking and infilling pretraining objectives, shifted labels in training, finding more compatible base models, and utilizing full bidirectional attention. In addition, we introduce a novel form of classifier guidance that leverages a reward model to further enhance generations at inference time. We find~\modelname~outperforms existing adapted diffusion models, highlighting the strength of our approach.
However, we also find that gaps still remain with AR models in some settings, leaving room for future research.

\section*{Limitations}

\paragraph{Sampling Speed} 


Although \modelname~is faster than its AR counterpart in basic generation settings, there are abundant rooms for improvement. In this work, we uniformly accelerate through reverse diffusion by using fewer backward steps than we do during training. However, incorporating recent work in computer vision for single step sampling in diffusion-based models~\citep{song2023consistency}, applying kernel or systems level optimizations~\cite{kwon2023efficient}, or investigating early stopping mechanisms as noted in Sec.~\ref{sec:predictions_evolve}, could be interesting avenues for further improvement.


\paragraph{Chat Performance Relative to AR Models} As noted in this paper, our best models still lag behind AR models on AlpacaEval and GSM8k when trained on Tulu 2 SFT data. As such, our model is a weaker generalist model than its AR counterparts. We hope that future improvements in data quality could help close the gap -- while Dolma is good quality data, we still find that continued pretraining of Mistral on Dolma results in degraded performance (Table~\ref{tab:instruction_performance}), suggesting that our adaptation data can be improved relative to the original Mistral pretraining data. Additionally, we expect a more thorough search and tuning of hyperparameters such as the learning rate (using decay instead of a constant schedule) to further improve performance. 

\section*{Ethics Statement}

In this work, we focused on improving an alternate modelling framework that diverges from popular language models. While we did not explicitly investigate this in our work, it is likely that our models display similar issues to their autoregressive counterparts when it comes to producing toxic and biased content~\citep{lmharms, sheng-etal-2021-societal}. However, we hope that the inherent controllability of the diffusion framework~\citep{li2022diffusion} may allow greater ability to reduce and avoid such harms. Examining how results around toxic and harmful generations of autoregressive setups transfer to diffusion models remains an open area for future investigation and improvement.

\section*{Acknowledgements}

We express gratitude to Matthew Peters for help and advice during earlier phases of this project. We also thank members of UW NLP for feedback. 
This research was supported in part by compute credits from Google.

\bibliography{anthology,refs}

\newpage
\section*{Appendix}

\appendix

\section{Adaptation Training Details}
\label{app:further_training}

\subsection{Pretraining Masking}

For pretraining, we use a UL2-inspired masking scheme~\citep{tay2023ul}, which involves randomly sampling different masking strategies. For each batch, we randomly select (with equal weighting) \textbf{T5}, 
\textbf{aggressive T5}, or \textbf{prefix-LM} masking. We detail each masking strategy below:

\begin{itemize}
    \item \textbf{T5}: This follows the T5 masking strategy~\citep{2020t5}, in which non-overlapping mask spans consisting of 3 or 8 tokens are applied until roughly 15\% of the input sequence is masked.
    \item \textbf{Aggressive T5}: This is a more aggressive variant of the T5 strategy above. We apply masking spans consisting of 3, 8, or 48 tokens until roughly 50\% of the input sequence is masked.
    \item \textbf{Prefix-LM}: This follows the prefix-LM objective~\citep{j2018generating}, where only the final tokens in the input sequence are masked, better resembling downstream usage (where we prompt an LM and want to generate a completion). We simply mask the first half of the input sequence and train the model to predict the rest.
\end{itemize}

Note that even with UL2, our training objective is the usual cross-entropy loss, and there are no modifications required: we simply take the model predictions for all output positions and use those for cross-entropy loss. Concretely, we first mask out random spans from the input text according to UL2. Next, we add noise to produce noisy word embeddings. The model then has to predict the actual words given these inputs, where the softmax output is paired with the usual cross-entropy loss.


Intuitively, adding span masks (1) acts as a strong regularizer that makes token prediction more difficult, (2) prepares the model for more flexible generation, e.g., filling, and (3) more directly encourages the use of bidirectional context during generation.

Finally, we performed some early pilot experiments with masking ratios for UL2, primarily testing removing the “aggressive T5” option. We found that this underperformed the initial UL2 masking strategy in AlpacaEval, albeit using slightly different hyperparameters and training data than used in the main text. As such, we did not remove the aggressive T5 masking.

\subsection{Infrastructure}

We train all models on a cluster of H100s, with all jobs running on at most 1 node (8 GPUs). During training, we apply common techniques used during autoregressive training to reduce training time and improve overall efficiency: fused AdamW, Flash Attention 2~\citep{dao2023flashattention2}, gradient checkpointing, and \texttt{bfloat16} training.

Running the diffusion adaptation training for 35,000 training steps (with sequence length 512) takes roughly 250 H100-hours, while running for 200,000 steps (with sequence length 2048) takes roughly 2,000 H100-hours. Based on commonly used training cost estimates from~\citet{kaplan2020scalinglawsneurallanguage}, the estimated total FLOPs is given by 
\begin{align*}
C
&\approx 6 N D \\
&= 6 \cdot (7 \times 10^9 \text{ params}) \cdot (45 \times 10^9\text{ tokens}) \\
&= 1.89 \times 10^{21} \text{ FLOPs}.
\end{align*}
Running instruction tuning (with sequence length 2048) takes roughly 280 H100 hours.

\section{Baseline Details}
\label{app:baselines}

We provide more details on baselines used for Table~\ref{tab:instruction_performance} below:

\begin{itemize}
    \item \textbf{DiffuLlama}~\citep{gong2024scalingdiffusionlm}: We finetune the publicly released checkpoint on Tulu 2 (and GSM8k) data, using code provided by the authors.\footnote{\url{https://github.com/HKUNLP/DiffuLLaMA}}
    \item \textbf{Flan-XLM-R-D XXL}~\citep{ye2023diffusionlanguagemodelsperform}: We directly evaluate the publicly released instruction tuned checkpoint using code provided by the authors.\footnote{\url{https://github.com/yegcjs/DiffusionLLM}}
    \item \textbf{Mistral v0.1 AR}: Mistral v0.1 7B directly finetuned on Tulu 2.
    \item \textbf{Mistral v0.1 AR w/ cont. pretrain}: Mistral v0.1 trained on Dolma 1.7 for the same number of steps and tokens as \modelname, then finetuned on the Tulu 2 dataset.
    \item \textbf{Mistral v0.3 AR}: Mistral v0.3 7B directly finetuned on Tulu 3 data.
\end{itemize}

\section{Reward Model Training}
\label{app:rm_training}

For reward model (RM) training, we use a standard setup following prior work~\citep{ouyang2022traininglanguagemodelsfollow,ivison2024unpacking}. 
Our RM is simply a causal decoder-only model with the LM head replaced with a regression head that predicts a score given a prompt and completion. We initialize our RM from a Mistral v0.1 7B model trained on the Tulu 2 SFT mixture (trained using the same hyperparameters as \citet{ivison2023camelschangingclimateenhancing}). We then use a dataset\footnote{\url{https://huggingface.co/datasets/weqweasdas/preference_dataset_mixture2_and_safe_pku}} used to to train state-of-the-art Mistral-based RMs on RewardBench~\citep{lambert2024rewardbench}.
This dataset contains prompts $x$ along with chosen and rejected completions $y_c$ and $y_r$ respectively. We then optimize the RM with
$$
\mathcal{L} = -\mathbb{E}_{(x, y_c, y_r) \sim \mathcal{D}} \big[ \log \sigma \big( R(x, y_c) - R(x, y_r) \big) \big],
\label{eq:reward_objective}
$$
where $R$ is the reward model and $D$ our overall dataset. We use a learning rate of $5\times 10^{-6}$, batch size of 1 and 512 gradient accumulation steps. We train for 1 epoch, linearly warming up the learning rate for the first 3\% of steps and linearly decaying to 0 over training afterwards. We use a maximum sample length of 2048 and filter out longer examples during training. The resulting model is used in our reward guidance experiments (Sec.~\ref{sec:reward_guidance}).

\section{Downstream Evaluation Details}
\label{app:eval_details}

Here we provide further details on the evaluation settings used for downstream and instruction-tuned evaluation:

\begin{itemize}
    \item \textbf{AlpacaEval}~\citep{alpaca_eval}: We use the package provided by \citet{alpaca_eval}, following the default setup for both AlpacaEval 1. We allow the evaluated model to generate up to 2048 tokens, without specifying special stop sequences. We use no few-shot samples for this evaluation. For (instruction-tuned) DiffuLlama, we found it useful to append `Response:' after the user instruction to obtain a response.
    \item \textbf{SQuAD}~\citep{rajpurkar-etal-2016-squad}: We randomly chose 512 samples from the SQuAD validation split as a test set. We chose 512 to reduce the computational cost of evaluation. We include the article containing the answer in the prompt, and include 3 in-context examples (randomly selected from the train set) in order to ensure the model outputs in the desired format. We report text-based F1.
    \item \textbf{GSM8k}~\citep{cobbe2021gsm8k}: We evaluate models on the full test set of GSM. Following \citet{wei2022chain}, we evaluate with chain-of-thought. We use 8 few-shot in-context examples. Because all answers in GSM are numbers, we extract the last number in the model response as the final answer. We report average accuracy across test examples. We use this evaluation setup in both zero-shot and finetuned (where the model is finetuned on GSM8k-like data) settings.
    \item \textbf{TriviaQA}~\citep{joshi-etal-2017-triviaqa}: We select 2000 examples from TriviaQA following \citet{gong2024scalingdiffusionlm}, and just prompt with the question alone. We use a 2-shot prompt with two examples from the train split in order to encourage the model to use the right format. We report exact match following \citet{gong2024scalingdiffusionlm}, where we give a prediction `1' if any answer alias appears in the model prediction (after minor post-processing) and `0' otherwise.
    \item \textbf{IFEval}~\citep{zhou2023instruction}: We use the entire IFEval suite, and report the loose prompt accuracy.
    \item \textbf{Big Bench Hard}~\citep{suzgun2022challenging, srivastava2022beyond}: We use the entire set of tasks and examples. If the prompt is too long for the model (as is the case for Flan-XLM-R-D XXL, which has a maximum context length of 512), we left-truncate the prompt to fit. We extract the answer using regular expresssions (searching for `The answer is X') and report exact match accuracy.
\end{itemize}

We use the Tulu chat template during evaluation for models that have undergone Tulu-based instruction tuning, and no chat template for other models.

\section{Sampling Speed}
\label{app:sampling_speed}

Recall that a single forward pass for a decoder-only transformer involves

$$
C_\mathrm{forward} \approx 2N + 2 n_\mathrm{layer} \ell d_\mathrm{model}
$$
add-multiply operations, where $N$ denotes the number of parameters; $n_\mathrm{layer}$, number of layers; $\ell$, input sequence length; and $d_\mathrm{model}$, hidden size~\cite{kaplan2020scalinglawsneurallanguage}. Thus, the full cost of generating a sequence of length $L$ is given by

\begin{align*}
C_\mathrm{AR} 
&\approx \sum_{\ell = 1}^L 2N + 2 n_\mathrm{layer} \ell d_\mathrm{model} \\
&= 2NL + n_\mathrm{layer} L (L + 1) d_\mathrm{model}.
\end{align*}

For diffusion, since we use the same transformer backbone, $C_\mathrm{forward}$ remains the same. However, instead of a varying $\ell$, we use a fixed-size sequence of length $L$ across all $B$ backward diffusion steps, which gives us

\begin{align*}
C_\mathrm{diffusion} 
&\approx B \times (2N + 2 n_\mathrm{layer} L d_\mathrm{model}) \\
&= 2NB + 2 n_\mathrm{layer} B L d_\mathrm{model}.
\end{align*}
We see that when $B < L + 1$, $C_\mathrm{diffusion}  < C_\mathrm{AR}$. In the~\modelname~ setup, $L = 2048$ and $B = 100$, so this relationship holds.

\section{Model Prediction Visualization}
\label{app:confidence_over_steps}

\begin{figure*}[!ht]
    \centering
    \includegraphics[width=\linewidth]{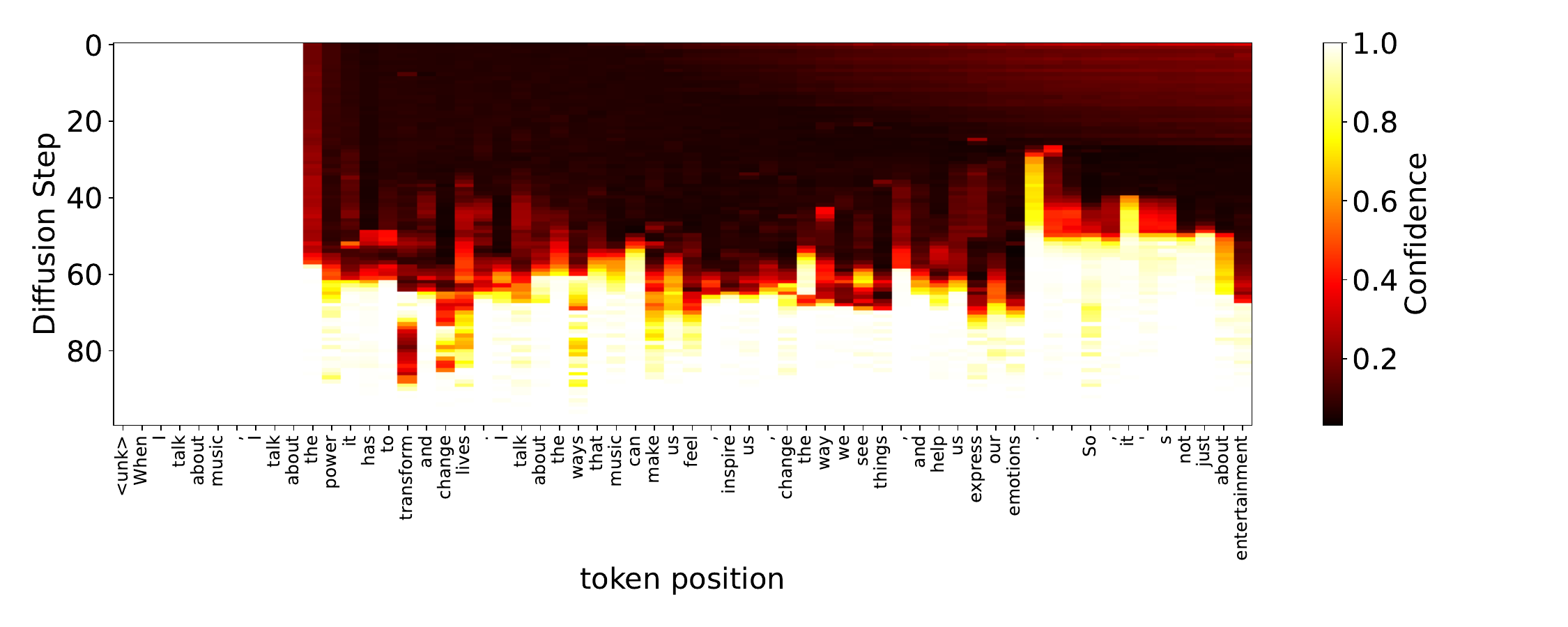}
    \caption{Confidence over diffusion steps for \modelname~with the prompt `When I talk about music, I.' Backward diffusion time flows from top to bottom. At about 60 diffusion steps, the sequence is more or less determined. Note we set confidence to 1 for prompt (leftmost) tokens.}
    \label{fig:confidence_over_steps}
\end{figure*}

In Figure~\ref{fig:confidence_over_steps}, we show how the confidence of the top-predicted token changes over diffusion steps when performing inference with \modelname.

\section{Example Reasoning Generations}
\label{app:reasoning_generations_app}

\begin{figure*}[t]
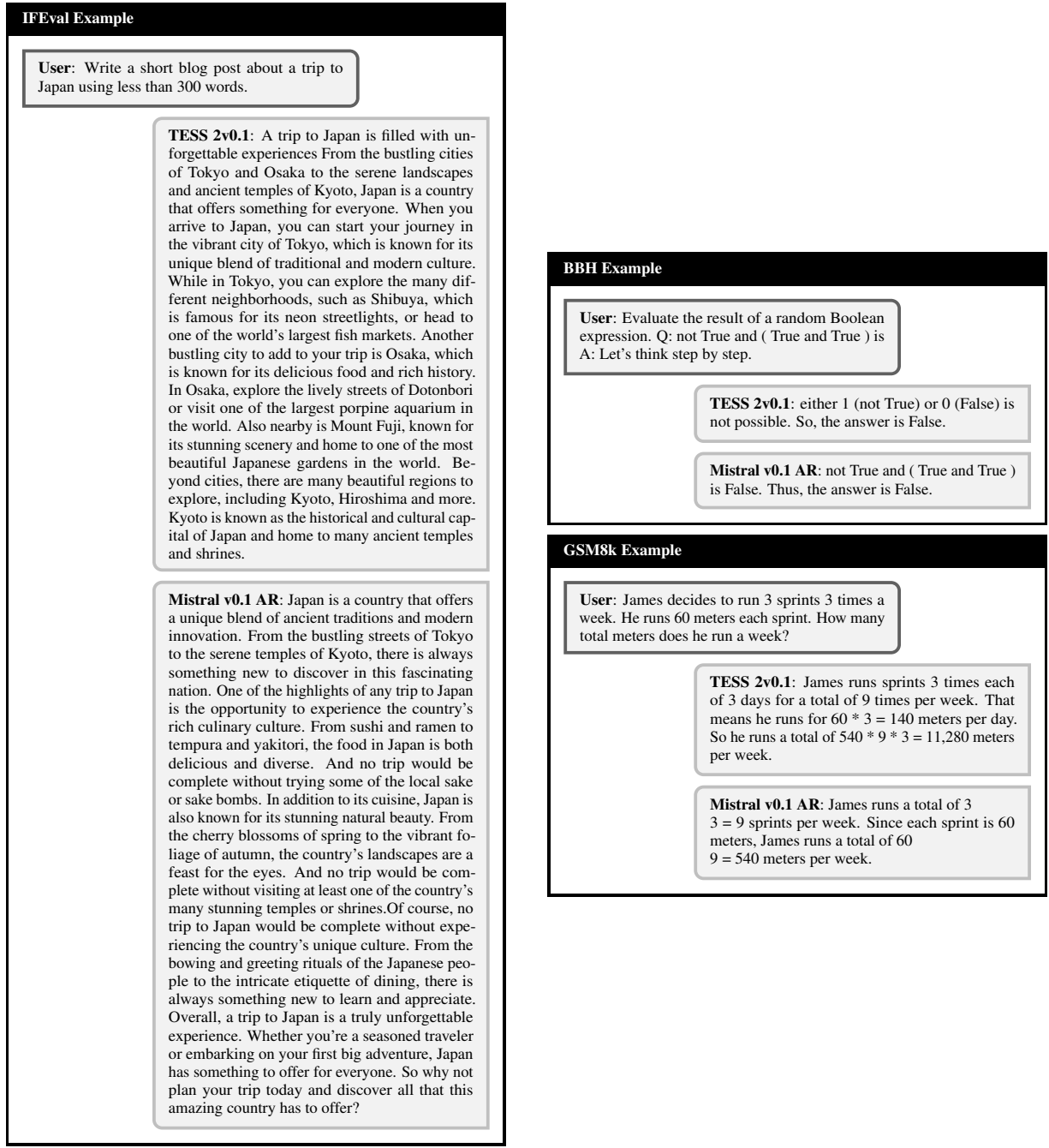

\centering
\begin{minipage}{0.48\textwidth}
\centering
\begin{subfigure}{\textwidth}
\centering
{\scriptsize
\newtcolorbox{compactbox}[2][]{colback=white!5!white, colframe=black!75!black, sharp corners,
    boxrule=0.5mm, boxsep=1mm, left=1mm, right=1mm, top=1mm, bottom=1mm,
    width=\linewidth, title=#2, fonttitle=\bfseries, #1}

\newtcolorbox{compactboxright}[2][]{ boxrule=0.5mm, boxsep=1mm, left=1mm, right=1mm, top=0.5mm, bottom=0.5mm, sharp corners=east, leftright skip=1cm,
    width=\dimexpr\textwidth-2cm\relax,
    enlarge left by=2cm,
    enlarge right by=0cm, title=#2, fonttitle=\bfseries, #1}

\newtcolorbox{compactboxleft}[2][]{ boxrule=0.5mm, boxsep=1mm, left=1mm, right=1mm,top=0.5mm, bottom=0.5mm,
    sharp corners=west,
    leftright skip=1cm,
    width=\dimexpr\textwidth-2cm\relax,
    enlarge left by=0cm,
    enlarge right by=0cm,
    title=#2, fonttitle=\bfseries, #1}

\begin{compactbox}{IFEval Example}
\begin{compactboxleft}[colframe=gray!75!black]{}
\textbf{User}: Write a short blog post about a trip to Japan using less than 300 words.
\end{compactboxleft}

\begin{compactboxright}[colframe=white!75!black]{}
\textbf{\modelname v0.1}: A trip to Japan is filled with unforgettable experiences From the bustling cities of Tokyo and Osaka to the serene landscapes and ancient temples of Kyoto, Japan is a country that offers something for everyone.
When you arrive to Japan, you can start your journey in the vibrant city of Tokyo, which is known for its unique blend of traditional and modern culture. While in Tokyo, you can explore the many different neighborhoods, such as Shibuya, which is famous for its neon streetlights, or head to one of the world's largest fish markets.
Another bustling city to add to your trip is Osaka, which is known for its delicious food and rich history. In Osaka, explore the lively streets of Dotonbori or visit one of the largest porpine aquarium in the world. Also nearby is Mount Fuji, known for its stunning scenery and home to one of the most beautiful Japanese gardens in the world.
Beyond cities, there are many beautiful regions to explore, including Kyoto, Hiroshima and more. Kyoto is known as the historical and cultural capital of Japan and home to many ancient temples and shrines.
\end{compactboxright}

\begin{compactboxright}[colframe=white!75!black]{}
\textbf{Mistral v0.1 AR}: Japan is a country that offers a unique blend of ancient traditions and modern innovation. From the bustling streets of Tokyo to the serene temples of Kyoto, there is always something new to discover in this fascinating nation. One of the highlights of any trip to Japan is the opportunity to experience the country's rich culinary culture. From sushi and ramen to tempura and yakitori, the food in Japan is both delicious and diverse. And no trip would be complete without trying some of the local sake or sake bombs. In addition to its cuisine, Japan is also known for its stunning natural beauty. From the cherry blossoms of spring to the vibrant foliage of autumn, the country's landscapes are a feast for the eyes. And no trip would be complete without visiting at least one of the country's many stunning temples or shrines.Of course, no trip to Japan would be complete without experiencing the country's unique culture. From the bowing and greeting rituals of the Japanese people to the intricate etiquette of dining, there is always something new to learn and appreciate. Overall, a trip to Japan is a truly unforgettable experience. Whether you're a seasoned traveler or embarking on your first big adventure, Japan has something to offer for everyone. So why not plan your trip today and discover all that this amazing country has to offer?
\end{compactboxright}
\end{compactbox}
}
\end{subfigure}
\end{minipage}
\hfill
\begin{minipage}{0.48\textwidth}
\centering
\begin{subfigure}{\textwidth}
\centering
{\scriptsize
\newtcolorbox{compactbox}[2][]{colback=white!5!white, colframe=black!75!black, sharp corners,
    boxrule=0.5mm, boxsep=1mm, left=1mm, right=1mm, top=1mm, bottom=1mm,
    width=\linewidth, title=#2, fonttitle=\bfseries, #1}

\newtcolorbox{compactboxright}[2][]{ boxrule=0.5mm, boxsep=1mm, left=1mm, right=1mm, top=0.5mm, bottom=0.5mm, sharp corners=east, leftright skip=1cm,
    width=\dimexpr\textwidth-2cm\relax,
    enlarge left by=2cm,
    enlarge right by=0cm, title=#2, fonttitle=\bfseries, #1}

\newtcolorbox{compactboxleft}[2][]{ boxrule=0.5mm, boxsep=1mm, left=1mm, right=1mm,top=0.5mm, bottom=0.5mm,
    sharp corners=west,
    leftright skip=1cm,
    width=\dimexpr\textwidth-2cm\relax,
    enlarge left by=0cm,
    enlarge right by=0cm,
    title=#2, fonttitle=\bfseries, #1}

\begin{compactbox}{BBH Example}
\begin{compactboxleft}[colframe=gray!75!black]{}
\textbf{User}: Evaluate the result of a random Boolean expression.
Q: not True and ( True and True ) is
A: Let's think step by step.
\end{compactboxleft}

\begin{compactboxright}[colframe=white!75!black]{}
\textbf{\modelname v0.1}: either 1 (not True) or 0 (False) is not possible. So, the answer is False.
\end{compactboxright}

\begin{compactboxright}[colframe=white!75!black]{}
\textbf{Mistral v0.1 AR}: not True and ( True and True ) is False. Thus, the answer is False.
\end{compactboxright}

\end{compactbox}

\begin{compactbox}{GSM8k Example}
\begin{compactboxleft}[colframe=gray!75!black]{}
\textbf{User}: James decides to run 3 sprints 3 times a week. He runs 60 meters each sprint. How many total meters does he run a week?
\end{compactboxleft}

\begin{compactboxright}[colframe=white!75!black]{}
\textbf{\modelname v0.1}: James runs sprints 3 times each of 3 days for a total of 9 times per week. That means he runs for 60 * 3 = 140 meters per day. So he runs a total of 540 * 9 * 3 = 11,280 meters per week.
\end{compactboxright}

\begin{compactboxright}[colframe=white!75!black]{}
\textbf{Mistral v0.1 AR}: James runs a total of 3 \\* 3 = 9 sprints per week. Since each sprint is 60 meters, James runs a total of 60 \\* 9 = 540 meters per week.
\end{compactboxright}

\end{compactbox}
}
\end{subfigure}
\end{minipage}
\caption{Generations from \modelname and Mistral AR v0.1. We find that \modelname~can produce long, coherent generations when prompted, but often makes reasoning mistakes, as seen in the GSM8k example.}
\label{fig:generation_examples_reasoning}
\end{figure*}

We compare \modelname and Mistral AR generations in Figure~\ref{fig:generation_examples_reasoning}. We see that while \modelname~can indeed produce long, coherent generations, it struggles with precise reasoning, as seen in its incorrect mathematical reasoning.

\section{Example Generations}

\begin{figure*}[t]
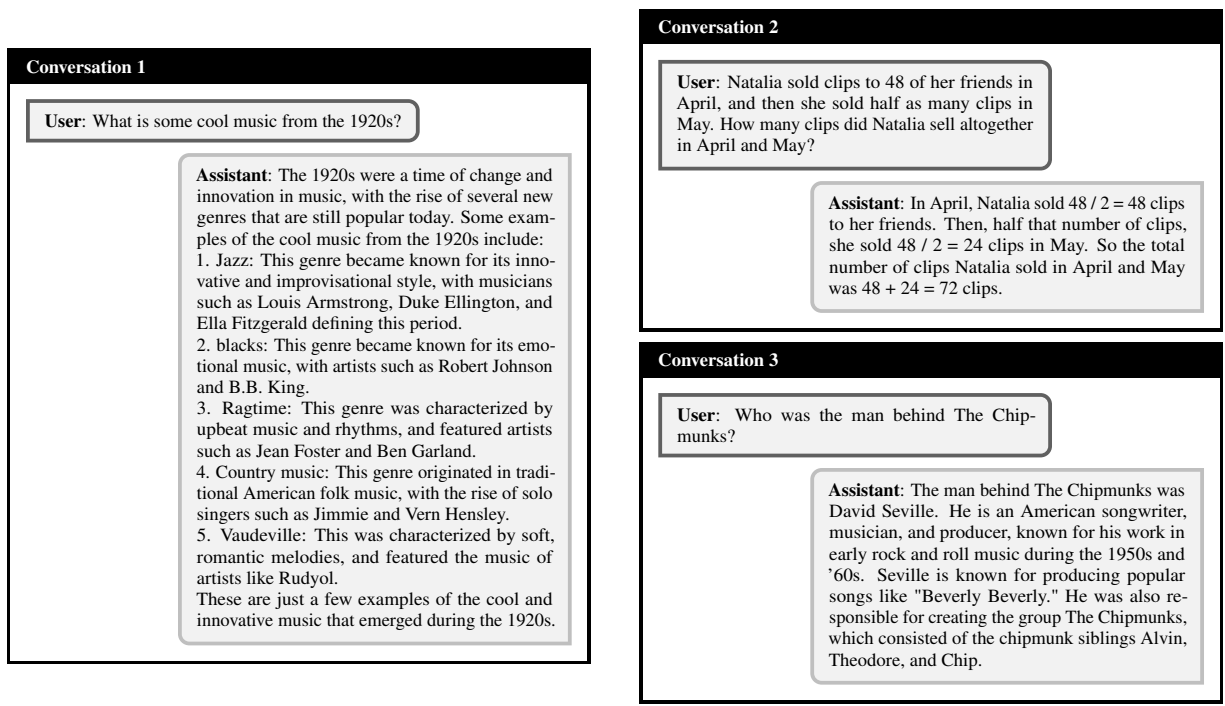

\centering
\begin{minipage}{0.48\textwidth}
\centering
\begin{subfigure}{\textwidth}
\centering
{\scriptsize
\newtcolorbox{compactbox}[2][]{colback=white!5!white, colframe=black!75!black, sharp corners,
    boxrule=0.5mm, boxsep=1mm, left=1mm, right=1mm, top=1mm, bottom=1mm,
    width=\linewidth, title=#2, fonttitle=\bfseries, #1}

\newtcolorbox{compactboxright}[2][]{ boxrule=0.5mm, boxsep=1mm, left=1mm, right=1mm, top=0.5mm, bottom=0.5mm, sharp corners=east, leftright skip=1cm,
    width=\dimexpr\textwidth-2cm\relax,
    enlarge left by=2cm,
    enlarge right by=0cm, title=#2, fonttitle=\bfseries, #1}

\newtcolorbox{compactboxleft}[2][]{ boxrule=0.5mm, boxsep=1mm, left=1mm, right=1mm,top=0.5mm, bottom=0.5mm,
    sharp corners=west,
    leftright skip=1cm,
    width=\dimexpr\textwidth-2cm\relax,
    enlarge left by=0cm,
    enlarge right by=0cm,
    title=#2, fonttitle=\bfseries, #1}

\begin{compactbox}{Conversation 1}
\begin{compactboxleft}[colframe=gray!75!black]{}
\textbf{User}: What is some cool music from the 1920s?
\end{compactboxleft}

\begin{compactboxright}[colframe=white!75!black]{}
\textbf{Assistant}: The 1920s were a time of change and innovation in music, with the rise of several new genres that are still popular today. Some examples of the cool music from the 1920s include: \\
1. Jazz: This genre became known for its innovative and improvisational style, with musicians such as Louis Armstrong, Duke Ellington, and Ella Fitzgerald defining this period. \\
2. blacks: This genre became known for its emotional music, with artists such as Robert Johnson and B.B. King. \\
3. Ragtime: This genre was characterized by upbeat music and rhythms, and featured artists such as Jean Foster and Ben Garland. \\
4. Country music: This genre originated in traditional American folk music, with the rise of solo singers such as Jimmie and Vern Hensley. \\
5. Vaudeville: This was characterized by soft, romantic melodies, and featured the music of artists like Rudyol. \\
These are just a few examples of the cool and innovative music that emerged during the 1920s.
\end{compactboxright}
\end{compactbox}
}
\end{subfigure}
\end{minipage}
\hfill
\begin{minipage}{0.48\textwidth}
\centering
\begin{subfigure}{\textwidth}
\centering
{\scriptsize
\newtcolorbox{compactbox}[2][]{colback=white!5!white, colframe=black!75!black, sharp corners,
    boxrule=0.5mm, boxsep=1mm, left=1mm, right=1mm, top=1mm, bottom=1mm,
    width=\linewidth, title=#2, fonttitle=\bfseries, #1}

\newtcolorbox{compactboxright}[2][]{ boxrule=0.5mm, boxsep=1mm, left=1mm, right=1mm, top=0.5mm, bottom=0.5mm, sharp corners=east, leftright skip=1cm,
    width=\dimexpr\textwidth-2cm\relax,
    enlarge left by=2cm,
    enlarge right by=0cm, title=#2, fonttitle=\bfseries, #1}

\newtcolorbox{compactboxleft}[2][]{ boxrule=0.5mm, boxsep=1mm, left=1mm, right=1mm,top=0.5mm, bottom=0.5mm,
    sharp corners=west,
    leftright skip=1cm,
    width=\dimexpr\textwidth-2cm\relax,
    enlarge left by=0cm,
    enlarge right by=0cm,
    title=#2, fonttitle=\bfseries, #1}

\begin{compactbox}{Conversation 2}
\begin{compactboxleft}[colframe=gray!75!black]{}
\textbf{User}: Natalia sold clips to 48 of her friends in April, and then she sold half as many clips in May. How many clips did Natalia sell altogether in April and May?
\end{compactboxleft}

\begin{compactboxright}[colframe=white!75!black]{}
\textbf{Assistant}: In April, Natalia sold 48 / 2 = 48 clips to her friends. Then, half that number of clips, she sold 48 / 2 = 24 clips in May. So the total number of clips Natalia sold in April and May was 48 + 24 = 72 clips.
\end{compactboxright}

\end{compactbox}

\begin{compactbox}{Conversation 3}

\begin{compactboxleft}[colframe=gray!75!black]{}
\textbf{User}: Who was the man behind The Chipmunks?
\end{compactboxleft}

\begin{compactboxright}[colframe=white!75!black]{}
\textbf{Assistant}: The man behind The Chipmunks was David Seville. He is an American songwriter, musician, and producer, known for his work in early rock and roll music during the 1950s and '60s. Seville is known for producing popular songs like "Beverly Beverly." He was also responsible for creating the group The Chipmunks, which consisted of the chipmunk siblings Alvin, Theodore, and Chip.
\end{compactboxright}
\end{compactbox}
}
\end{subfigure}
\end{minipage}
\caption{Non-cherry picked sample generations from \modelname. Note we use a shorter generation length (256 tokens) than during training and evaluation (2048 tokens) for ease of reading.}
\label{fig:generation_examples}
\end{figure*}

We provide some example generations from our model for more general prompts in Figure~\ref{fig:generation_examples}.

\end{document}